\documentclass[11pt,a4paper]{article}
\usepackage[hyphens]{url}
\usepackage[hyperref]{acl2021}
\usepackage{amssymb}

\usepackage{times}
\usepackage{latexsym}
\usepackage{graphicx}
\usepackage{multirow,multicol}
\usepackage{amsthm,amsmath,amssymb}
\usepackage{mathrsfs}
\usepackage{booktabs} % Added
\usepackage{dingbat} % Added
\usepackage{multirow} % Added
\usepackage[bottom]{footmisc}

\usepackage{microtype}

\aclfinalcopy % Uncomment this line for the final submission
\def\aclpaperid{165} %  Enter the acl Paper ID here

\title{IIE-NLP-Eyas at SemEval-2021 Task 4: Enhancing PLM for ReCAM with Special Tokens, Re-Ranking, Siamese Encoders and Back Translation}

\author{
  Yuqiang Xie, Luxi Xing, Wei Peng, Yue Hu\footnotemark[1] \\
  Institute of Information Engineering, Chinese Academy of Sciences, Beijing, China \\
  School of Cyber Security, University of Chinese Academy of Sciences, Beijing, China \\
  {\tt \{xieyuqiang,xingluxi,pengwei,huyue\}@iie.ac.cn}
}

\begin{document}
\maketitle

\renewcommand{\thefootnote}{\fnsymbol{footnote}} %将脚注符号设置为fnsymbol类型，即特殊符号表示
\footnotetext[1]{Corresponding author.}

\begin{abstract}
This paper introduces our systems for all three subtasks of SemEval-2021 Task 4: Reading Comprehension of Abstract Meaning.
To help our model better represent and understand abstract concepts in natural language, we well-design many simple and effective approaches adapted to the backbone model (RoBERTa).
Specifically, we formalize the subtasks into the multiple-choice question answering format and add special tokens to abstract concepts, then, the final prediction of question answering is considered as the result of subtasks. Additionally, we employ many finetuning tricks to improve the performance.
Experimental results show that our approaches achieve significant performance compared with the baseline systems.
Our approaches achieve eighth rank on subtask-1 and tenth rank on subtask-2.
\end{abstract}

\section{Introduction}

The computer's ability in understanding, representing, and expressing abstract meaning is a fundamental problem towards achieving true natural language understanding. SemEval-2021 Task 4: Reading Comprehension of Abstract Meaning (ReCAM) provides a well-formed benchmark that aims to study the machine's ability in representing and understanding abstract concepts \cite{zheng-2021-semeval-task4}.

The Reading Comprehension of Abstract Meaning (ReCAM) task is divided into three subtasks, including Imperceptibility, Nonspecificility, and Interaction. Please refer to the task description paper \cite{zheng-2021-semeval-task4} for more details.
To address the above challenges in ReCAM, we first formalize all subtasks as a type of multiple-choice Question Answering (QA) task like \cite{Xing2020IIENLPNUTAS}. Recently, the the large Pre-trained Language Models (PLMs), such as GPT-2 \cite{Radford2019LanguageMA}, BERT \cite{DBLP:conf/naacl/DevlinCLT19}, RoBERTa \cite{DBLP:journals/corr/abs-1907-11692}, demonstrate its excellent ability in various natural language understanding tasks \cite{wang-etal-2018-glue,zellers-etal-2018-swag,zellers-etal-2019-hellaswag}. So, we employ the state-of-the-art PLM, RoBERTa, as our backbone model. Moreover, we well-design many simple and effective approaches to improve the performance of the backbone model, such as adding special tokens, sentence re-ranking, label smoothing and back translation.

This paper describes approaches for all subtasks developed by the IIE-NLP-Eyas Team (Natural Language Processing group of Institute of Information Engineering of the Chinese Academy of Sciences).
Our contributions are summarized as the followings:
\begin{itemize}
\item We design many simple and effective approaches to improve the performance of the PLMs on all three subtasks, such as adding special tokens, sentence re-ranking and so on;
\item Experiments demonstrate that the proposed methods achieve significant improvement compared with the PLMs baseline and we obtain the eighth-place in subtask-1 and the tenth-place in subtask-2 on the final official evaluation.
\end{itemize}

\section{Approaches}
\label{sect:approach}

Since the format of the tasks in ReCAM is the same, we use the unified framework to address all tasks. The following is the detail of our methods.

\paragraph{Task Definition}
We first present the description of symbols. Formally, suppose there are seven key elements in all subtasks, i.e. $\{D, Q, A_1, A_2, A_3, A_4, A_5\}$.
We suppose the $D$ denotes the given article, the $Q$ denotes the summary of the article with a placeholder, the $A_*$ denotes the candidate abstract concepts for all subtasks to fill in the placeholder.

\paragraph{Multi-Choice Based Model}
The pre-trained language models have made a great contribution to MRC tasks. Recently, a significant milestone is the BERT \cite{DBLP:conf/naacl/DevlinCLT19}, which gets new state-of-the-art results on eleven natural language processing tasks.
In this section, we present the description of the multi-choice based model which we use in all subtasks. Consider the BERT-style model RoBERTa's \cite{DBLP:journals/corr/abs-1907-11692} stronger performance than BERT, we utilize it as our backbone model, which introduces more data and bigger models for better performance.
A multiple-choice based QA model $\mathcal{M}$ consists of a PLM encoder and a task-specific classification layer which includes a feed-forward neural network $f(\cdot)$ and a softmax operation.
For each pair of question-answer, the calculation of $\mathcal{M}$ is as follow:
\begin{eqnarray}
  score_i &=& \frac{exp(f(S_i))}{\sum_{i^\prime} exp(f(S_{i^\prime}))}\\
  S_i &=& \mbox{PLM}([Q;A_i;D])
\label{eq:basic-mc}
\end{eqnarray}
where the $[\cdot]$ is the input constructed according to the instruction of PLMs, and the $S_*$ is the final hidden state of the first token (\texttt{<s>}).
For more details, we refer to the original work of PLMs \cite{DBLP:journals/corr/abs-1907-11692}.
The candidate answer which owns a higher $score$ will be identified as the final prediction.
The model $\mathcal{M}$ is trained end-to-end with the cross-entropy objective function.

\paragraph{Special Tokens}
To help the model to the PLMs represent and understanding the abstract concept in textual descriptions, we add special tokens to enhance the semantic representation of candidate concepts. The idea is similar to the prompt template of  \cite{Xing2020IIENLPNUTAS}. We use \texttt{<e>} and \texttt{</e>} to add on both ends of the abstract concept, i.e. \texttt{<e> abstract concept </e>}. We also tried many other special tokens which will be discussed in section \ref{sect-analysis}.

\paragraph{Sentence Ranking}
As the given passage is too long to be deal with the Pre-trained Language Models (PLMs), we consider refining the passage input by rearranging the order of the sentences in the passage.
With this reorder process, the sentence, which is more critical to the question, can appear at the beginning of the passage.
Although the passage's sequential information is sacrificed, we keep the more question-relevant information of the passage.
Supposing the Passage $D$ contains $N$ sentences, i.e., $D=\{W_1, W_2, ..., W_N\}$, where each sentence $W_n = \{t_1, t_2, ..., t_M\}$ including $M$ tokens.
We denote the given cloze-style question as $Q$.
To rank the sentences in $D$, we resort BERT to compute the similarity score between each sentence, i.e. $W_n$, and $Q$ following the algorithm in \citet{bert-score}.
After ranking, the sentences in $D$ are sorted in descending order of similarity scores, and we can get a rearranged passage $\hat{D}$ as the passage input to the QA model.

\paragraph{Siamese Encoders}

When exploring the dataset, we find that the complete question statement, representing the result statement after replacing the placeholder token with the candidate option, also contains the semantic information which can help to make the judgment about options.
Based on the observation, we propose a siamese encoders based architecture to inject the additional complete question statement information while not influence the input with passage.
On the other hand, it can be seen as introducing an auxiliary task to assist the main task.
Specifically, the training of siamese encoder based architecture is as following:
\begin{eqnarray}
 l^1_i &=& PLM([\hat{Q}_i])[0] \\
 l^2_i &=& PLM([Q;A_i;D])[0] \\
 P^1(A_i|\hat{Q}) &=& \mbox{softmax}(f(l^1_i)) \\
 P^2(A_i|D,Q) &=& \mbox{softmax}(f(l^2_i))
\end{eqnarray}
where the $PLM(\cdot)$ stands for PLM encoder, $\hat{Q}_i$ is the complete question statement, $i$ indicates the i-th candidate answer, $f(\cdot)$ is the feed forward network.
To coordinate the two losses, we opt for an uncertainty loss \cite{DBLP:conf/cvpr/KendallGC18} to adjust it adaptively through $\sigma_{\{1,2\}}$ as: $ \mathcal{L}(\theta, \sigma_1, \sigma_2) = \frac{1}{2\sigma_1^2} \mathcal{L}^1(\theta) + \frac{1}{2\sigma_2^2} \mathcal{L}^2(\theta) + \mbox{log} \sigma_1^2 \sigma_2^2$, where $\mathcal{L}^{\{1,2\}}$ are the cross-entropy loss between the model prediction $P^{\{1,2\}}$ and the ground truth label respectively.

\paragraph{Back Translation}
Generally speaking, more successful neural networks require a large number of parameters, often in the millions. In order to make the neural network implements correctly, a lot of data is needed for training, but in actual situations, there is not as much data as we thought. The role of data augment includes two aspects. One is to increase the amount of training data and improve the generalization ability of the model. The other is to increase the noise data and improve the robustness of the model. A large number of the works \cite{DBLP:journals/corr/abs-1809-06839,DBLP:journals/bioinformatics/BloiceRH19,DBLP:journals/corr/abs-2001-04086,DBLP:conf/cvpr/CubukZSL20,DBLP:conf/ijcai/SatoSS018,DBLP:conf/iclr/ZhuCGSGL20} consider the data augment to make better performances. In the field of computer vision, a lot of work \cite{DBLP:journals/corr/abs-1809-06839,DBLP:journals/bioinformatics/BloiceRH19,DBLP:journals/corr/abs-2001-04086,DBLP:conf/cvpr/CubukZSL20} uses existing data to perform operations, such as flipping, translation or rotation, to create more data, so that neural networks have better generalization effects. Adding Gaussian distribution to text processing \cite{DBLP:conf/ijcai/SatoSS018} can also achieve the effect of data augment. Besides, some works \cite{DBLP:conf/iclr/MiyatoDG17,DBLP:conf/iclr/ZhuCGSGL20} utilize the adversarial training methods to do the data augment. For convenience and simplicity, we adopt the back translation \cite{DBLP:conf/acl/SennrichHB16} to increase the amount of training data, which used to construct pseudo parallel corpus in unsupervised machine translation \cite{DBLP:conf/iclr/LampleCDR18}. Specifically, we use the Google API\footnote[2]{The web page is available at \small{~https://translate.google.com}} to translate the passage into French, and then translate the translation into English in turn. The pseudo parallel corpus can be obtained as:
\begin{eqnarray}
\{D'\} &=& bkt(\{D\})
\label{eq:basic-bk}
\end{eqnarray}
where $\{D'\}$ means the translated English corpus that we used as data agument, $bkt$ is back translation.

As for the question, given the existence of the special character \textit{placeholder}, forced translation may result in grammatical errors and semantic gaps. Therefore, the questions and options will be kept original. After getting the pseudo parallel corpus, we train our model with the training data together with the cross-entropy loss function.

\begin{table}[t!]
\centering
\resizebox{0.95\linewidth}{!}{
\begin{tabular}{lcccc}
\toprule[1pt]
Subtask                  & Train  & Trail & Dev   & Test   \\
\toprule[0.5pt]
Imperceptibility        & 3227  & 1000 & 837   & 2025 \\
Nonspecificility        & 3318  & 1000 & 851   & 2017 \\
\toprule[1pt]
\end{tabular}}
\caption{Data scale of each subtask.}
\label{data}
\end{table}

\begin{table}[t!]
\centering
\resizebox{0.95\linewidth}{!}{
\begin{tabular}{lc}
\toprule[1pt]
Hyper-parameter                  & Value      \\
\toprule[0.5pt]
LR                               & \{1e-5, 2e-5\} \\
Batch size                       & \{16, 32\}    \\
Gradient norm                    & 1.0        \\
Warm-up                          & \{0.1, 1, 2\}        \\
Max. input length (\# subwords)  & 200        \\
Epochs                    & [3, 10] \\
\toprule[1pt]
\end{tabular}}
\caption{Hyper-parameters of our approach.}
\label{parameter}
\end{table}

\paragraph{Label Smoothing}

Furthermore, for improving the generalization ability of the model trained on sole task and prevent the overconfidence of model, we consider training model with label smoothing \cite{DBLP:journals/tsp/MillerRRG96,DBLP:conf/iclr/PereyraTCKH17}.
When training with label smoothing, the hard one-hot label distribution is replaced with a softened label distribution through a smoothing value $\alpha$, which is a hyper-parameter.
In our experiments, we set the smoothing value $\alpha = 0.1$.

\section{Experiments and Results}

\begin{table}[t]
\centering
\resizebox{1\linewidth}{!}{
\begin{tabular}{lcc}
\toprule[1pt]
Models & Trial Acc. & Dev Acc.  \\
\toprule[0.5pt]
\textsc{Roberta}$_{\mbox{\scriptsize LARGE}}$\cite{DBLP:journals/corr/abs-1907-11692} & 85.85 & 82.12  \\
(1) w/ special tokens   & \bf 87.81  & \bf 87.69 \\
(2) w/ sentence ranking  & 86.54  &  83.52 \\
(3) w/ label smoothing   & 86.88  &  85.85  \\
(4) w/ siamese encoders   & 86.62  & 83.22   \\
(5) w/ back translation   & 87.23  & 84.32 \\
\toprule[0.5pt]
Our Approach & \bf 87.81  & \bf 87.69 \\
\toprule[1pt]
\end{tabular}}
\caption{The results of our approach on subtask-1. Our approach is the final, stable and best model: \textsc{Roberta}$_{\mbox{\scriptsize LARGE}}$ with special tokens.}
\label{result-1}
\end{table}

\subsection{Experimental Setup}
In all subtasks, the scale of each task is shown in Table \ref{data}. We train the model on training data and the related pseudo data generated by back translation, then select hyper-parameters based on the best performing model on the dev set, and then report results on the test set.

Our system is implemented with PyTorch and we use the PyTorch version of the pre-trained language models\footnote[3]{https://github.com/huggingface/transformers}.
We employ RoBERTa \cite{DBLP:journals/corr/abs-1907-11692} large model as our PLM encoder in Equation \ref{eq:basic-mc}.
The Adam optimizer \cite{kingma2014adam} is used to fine-tune the model.
We introduce the detailed setup of the best model on the development dataset.
For subtask-1 and subtask-2, the hyper-parameters are shown in Table \ref{parameter}.

\subsection{Evaluation Results}
\label{sect-sub:exp-result}

\paragraph{Imperceptibility}
From Table \ref{result-1}, we can see the results of our approach on subtask-1 of ReCAM. Compared to the backbone model RoBERTa large model, our methods achieve significant improvements. It is interesting that the special token is the most helpful part for the Imperceptibility subtask.

\paragraph{Nonspecificility}
Table \ref{result-2} shows the results of our approach on subtask-2 of ReCAM. Similarly, the models with special tokens work well on the Nonspecificility subtask. Compared to the backbone model RoBERTa large model, our methods achieve better improvements.

\begin{table}[t]
\centering
\resizebox{1\linewidth}{!}{
\begin{tabular}{lcc}
\toprule[1pt]
Models & Trial Acc. & Dev Acc.  \\
\toprule[0.5pt]
\textsc{Roberta}$_{\mbox{\scriptsize LARGE}}$\cite{DBLP:journals/corr/abs-1907-11692} & \textbf{88.51} &  85.93  \\
(1) w/ special tokens   & 87.47  & 88.98 \\
(2) w/ sentence ranking  & 87.29  & 86.84 \\
(3) w/ label smoothing   & 87.67  & 87.08 \\
(4) w/ siamese encoders   & 87.34  & 86.18 \\
(5) w/ back translation   & 88.41  & 87.54 \\
\toprule[0.5pt]
Our Approach & 87.10  & \bf 89.54 \\
\toprule[1pt]
\end{tabular}}
\caption{The results of our approach on subtask-2. Our approach is the final, stable and best model: \textsc{Roberta}$_{\mbox{\scriptsize LARGE}}$ with special tokens and label smoothing.}
\label{result-2}
\end{table}

\paragraph{Interaction}

We also perform subtask-3 of ReCAM, Interaction, which aims to provide more insights into the relationship of the two views on abstractness. In this task, we test the performance of our system that is trained on one definition and evaluated on the other. The results of our system's performance on Imperceptibility and Nonspecificility subtasks which is shown in Table \ref{result-3}. We can find that our model is relatively robust for different abstract concepts.

\begin{table}[t]
\centering
\begin{tabular}{ccc}
\toprule[1pt]
Trained on & Tested on &  Test Acc. \\
\toprule[0.5pt]
Subtask-1 & Subtask-1  & 87.51 \\
Subtask-1 & Subtask-2  & 84.13 \\
Subtask-2 & Subtask-2  & 89.64 \\
Subtask-2 & Subtask-1  & 81.09 \\
\toprule[1pt]
\end{tabular}
\caption{The results of our approach on subtask-3.}
\label{result-3}
\end{table}

\begin{table}[t]
\centering
\begin{tabular}{lcc}
\toprule[1pt]
Special Token & Trial Acc. & Dev Acc. \\
\toprule[0.5pt]
\texttt{<e> </e>} & 88.01 & \bf 87.10 \\
\texttt{<\#> </\#>} & \bf 88.63 & 86.93 \\
\texttt{<\$> </\$>} & 88.12 & 86.26 \\
\texttt{\# /\#} & 87.34 & 85.89 \\
\texttt{\$ /\$} & 87.73 & 86.13 \\
\toprule[0.5pt]
N/A & 86.23 & 83.12 \\
\toprule[1pt]
\end{tabular}
\caption{The results of models with different special tokens on subtask-1.}
\label{spc}
\end{table}

\section{Analysis and Discussion}
\label{sect-analysis}

\subsection{Ablation Study}
In this part, we perform an ablation study of our approach. As shown in Table \ref{result-1} and \ref{result-2}, our proposed methods help the backbone model better represent and understand the abstract concepts. Note that the special tokens bring the PLMs with the best improvements in both subtask-1 and subtask-2. It is possible that the special tokens teach the model to focus on the abstract concept in a stronger manner. Moreover, other common tricks bring with little improvements.

\subsection{Discussion of Special Tokens}

We also search for the best special tokens for ReCAM on the dev set of subtask-1. Tabel \ref{spc} shows that \texttt{<e> </e>} enhance the representations of abstract concepts well.
Besides, the \texttt{<>} and \texttt{</>} could be helpful for PLMs to pay attention to the abstract concepts.
Moreover, it is interesting that each special token helps PLMs choose the right abstract concepts which submerged in long sequential tokens (including article and summary). This result strength that special tokens can enhance the representation of abstract concepts in PLM based approaches.

\section{Conclusion}
\label{sect-conclusion}
In this paper, we design many simple and effective approaches to improve the performance of the PLMs on all three subtasks.
Experiments demonstrate that the proposed methods achieve significant improvement compared with the PLMs baseline and we obtain the eighth-place in subtask-1 and tenth-place in subtask-2 on the final official evaluation. Moreover, we show that special tokens work well in enhancing PLMs for representating and understanding abstract concepts.

% \section*{Acknowledgments}

% The acknowledgments should go immediately before the references. Do not number the acknowledgments section.
% \textbf{Do not include this section when submitting your paper for review.}

\bibliographystyle{acl_natbib}
\bibliography{acl2021}

%\appendix

\end{document}